\documentclass{article}
\usepackage{spconf,amsmath,graphicx}
\usepackage{color}

\usepackage{tikz}
\usepackage{textcomp}
\usepackage{xcolor}
\usepackage{blindtext}
\usepackage{amssymb}
\usepackage{graphicx}
\usepackage{amsmath}
\usepackage{amstext}
\usepackage{subfigure}
\usepackage{enumitem}
\usepackage{url}
\usepackage{array}
\newcolumntype{x}[1]{>{\centering\arraybackslash\hspace{0pt}}p{#1}}
\usepackage{multirow}
\usepackage{bbding}
\usepackage{subfig}
\usepackage[T1]{fontenc}
\usepackage{graphics}
\usepackage{caption}

\usepackage{mwe} 
\usepackage{fancyhdr}
\usepackage{lipsum}
\title{Self-Attention Equipped Graph Convolutions for Disease Prediction} 
\name{Anees Kazi$^{*1}$, S.Arvind krishna$^{2}$, Shayan Shekarforoush$^{3}$, Karsten Kortuem$^{4}$, Shadi Albarqouni$^{1}$, Nassir Navab$^{1,5}$}
\address{$^{1}$Computer Aided Medical Procedures, Technische Universit\"at M\"unchen, Germany\\
$^{2}$National Institute of Technology Tiruchirappalli, India\\
$^{3}$Sharif University of Technology, Iran\\
$^{4}$Augenklinik der Universit\"at, Klinikum der Universit\"at M\"unchen, Germany\\
$^{5}$Johns Hopkins University, Baltimore MD, USA}

\begin{document}
%\ninept
%
\maketitle
\begin{abstract}
Multi-modal data comprising imaging (MRI, fMRI, PET, etc.) and non-imaging (clinical test, demographics, etc.) data can be collected together and used for disease prediction. Such diverse data gives complementary information about the patient's condition to make an informed diagnosis. A model capable of leveraging the individuality of each multi-modal data is required for better disease prediction. We propose a graph convolution based deep model which takes into account the distinctiveness of each element of the multi-modal data. We incorporate a novel self-attention layer, which weights every element of the demographic data by exploring its relation to the underlying disease. We demonstrate the superiority of our developed technique in terms of computational speed and performance when compared to state-of-the-art methods. Our method outperforms other methods with a significant margin.
\end{abstract}
\vspace{-0.4cm}
\section{Introduction}
\label{sec:intro}
Experts look at all the varied multi-modal data collected by imaging sources and non-imaging demographics (age, gender, weight, body-mass index) to take an informed decision for disease diagnosis. Such rich data is also exploited in Computer Aided Diagnosis systems (CADs) as complementary information. Current CAD systems combine all the complementary features by using
%regressing them \cite{lorenzi2016multimodal} or 
feature selection \cite{memarian2015multimodal}, or by reducing the dimensionality with  an autoencoder \cite{calhoun2016multimodal,tiwari2011multi,ngiam2011multimodal}. Works are also done with simply concatenating all the features to use deep learning based models \cite{Xu2016}.
%Modified Restricted Boltzmann Machine \cite{suk2014hierarchical}. 
All the above methods exploit the complementary information from available modalities at a global level but fail to optimally combine the varied information. For instance, the learned features are biased towards the single modality with dominant features and do not exploit the individuality of each modality. On top of that, each demographic information carries different relevance for the diagnosis of a disease. A model is required which is capable of evaluating the significance of every element of the demographic data and performing the prediction task based on the selective and weighted procedure for elements of demographic data. Such a scheme will boost the disease prediction task to incorporate more clinical semantics. %Graph regularization techniques are used to bring in the inventive notion of relationship between the subjects into the learning procedure \cite{albarqouni2015multi}.
\vspace{-0.1cm} 
\par Graphs provide a more such a way of using multi-modal data \cite{parisot2017spectral,kipf2016semi}. These methods leverage the similarities between subjects in terms of an affinity graph in the training process itself. 
% These methods focus more on the association between the patients with respect to either of the modalities and then solve the tasks such as disease prediction with features from other modalities. 
Most recent work \cite{parisot2017spectral} by presents an intelligent and novel use case of Graph Convolutional Networks (GCN) for the binary classification task. This allows convolutions to be used on graph-structured data, where each patient represents a node in the population level graph. %suffering from Autism Spectrum Disorders (ASD).
The method proposes to use each demographic information separately to construct a neighborhood graph. They eventually combine all the neighborhood graphs to get the average affinity graph, unlike the conventional methods, which fuses the information for the prediction task. This method, however, yields varied results for distinct input neighborhood graphs. Each of these affinity graphs and indirectly each element of the demographic data carries distinct neighborhood relationships (based on element dependent criteria) and statistical properties with respect to the entire population.
\vspace{-0.1cm} 
\par Our motivation is to analyze the impact and relevance of the neighborhood definitions on the final task of disease prediction. In addition to that, we want to investigate whether the relative weighting of meta-data can be automated. \textbf{Contributions:} 1) We propose a model capable of incorporating the information of each graph separately, 2) our design architecture bears a parallel setting of Graph Convolutional (GC) layers 3) we introduce a 'Self-Attention layer' which automatically learns the weighting for each meta-data with respect to its relevance to the prediction task, and 4) Our model outperforms the state-of-the-art method.\\

\vspace{-0.7cm} 
\section{Methodology}
\label{sec:format}
\begin{figure*}[t]
	\begin{center}
	\includegraphics[width=0.9\linewidth,height=0.25\textheight]{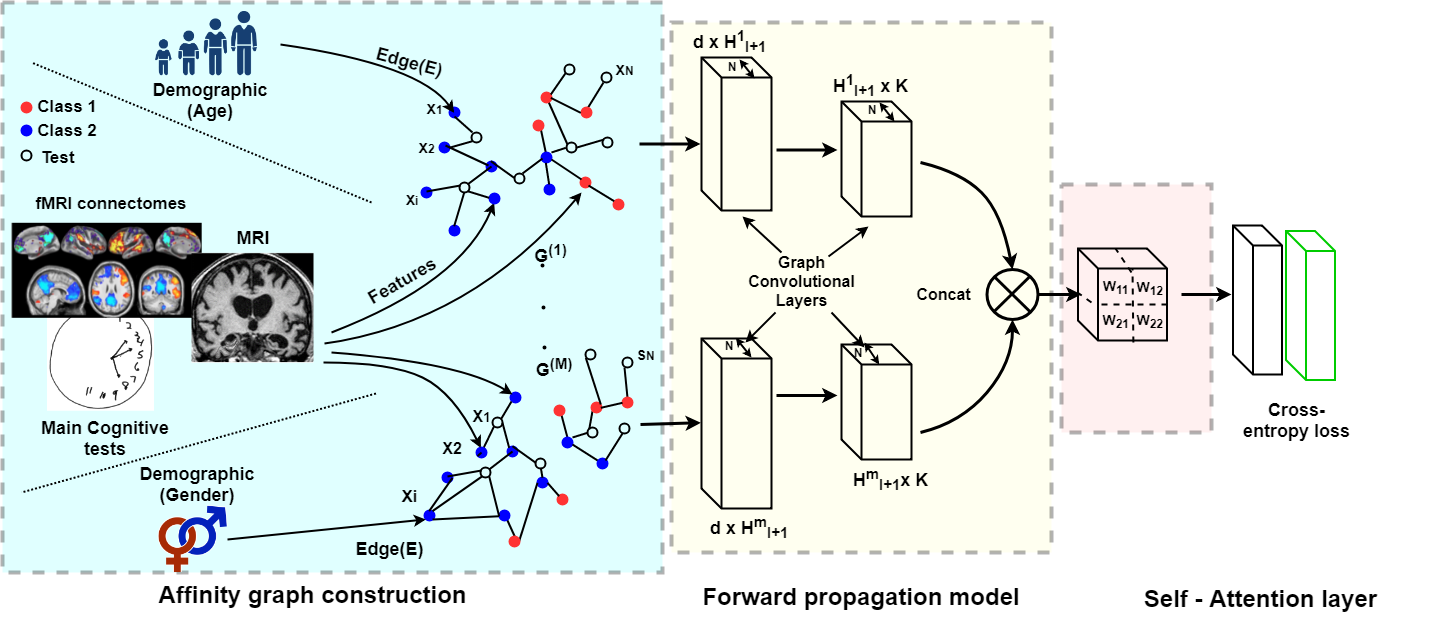}
	\caption{Figure describes the Multi-Layered Parallel Graph Convolutional Network with $M$=2. Two branches have same input features but input affinity matrix.}
	\label{fig:framework}
	\end{center}
\end{figure*}
% In this section we first describe the scenario and detailed notation, followed by  mathematical background of GCN with details of construction of affinity matrices $W^{(i)}$ and finally propagation model of proposed multi-layer graph network leveraging the defined affinity matrix and GCN.
%\vspace{-0.2cm} 
%\par Considering a database of $N$ subjects, each with data tuple of $d$ dimensional feature vector and $m$ elements from corresponding demographic data. Target is to classify the population in $K$ classes with an automatic weighting of demographic elements. The proposed model incorporates $M$ parallel branches leveraging one of the demographic element separately to learn the filters of each branch. Parallel branches consist of 2 Graph Convolutional layers each, which are finally fused using the self-attention  layer. Each branch receives common features for each node. Below we explain the initial setting. Mathematical background on GCN explains the propagation modal and labeling. We provide details on affinity graph construction and finally self-attention layer. 
%\textbf{\textit{Notation:}}

%Given a database $\mathcal{D} =\{X, Y\}$ that comprises $N$ subjects $X \in \mathbb{R}^{N\times d}$, each represented by a $d$-dimensional feature vector, and a subset of labeled subjects ($K$-classes) $Y_L \in \mathbb{R}^{L \times K}$, $K$ being one hot vector for our case and $M$ affinity graphs $G^{(m)} \in \mathbb{R}^{N \times N}$ computed from the respective demographic element. The task is to classify the nodes (patients) with unknown labels (test labels) to $K$ classes.
Given a dataset $\mathcal{D} =\{X, Y, \delta\}$ with $X \in \mathbb{R}^{N\times d}$ representing the feature matrix for $N$ patients and each one is provided with $d$-dimensional features. $Y$ represents the corresponding label matrix and $\delta$ the demographic data matrix. The task is to predict the class label $\hat{Y}$ for test subjects for $K$ classes. 
%Given a database $\mathcal{D} =\{X, Y, \delta\}$ with 
%$X \in \mathbb{R}^{N\times d}$ comprising $N$ subjects, each represented by a $d$-dimensional feature vector. $Y \in \mathbb{R}^{L \times K}$ available only of $L$ patients with $L \subset N$ and $K$ classes with one hot encoded representation .
%with $K$ being the number of classes and a one hot vector for our case and $L \subset N$ used as training samples and $N-L$ used for testing.
$\delta \in \mathbb{R}^{N\times M}$ represents that for each patient $M$-dimensional demographic data is provided. 
%$\delta \in \mathbb{R}^{N\times M}$ being demographic the $M$-dimensional data provided for each patient. 
The $m^{th}$ affinity graphs $G^{(m)} \in \mathbb{R}^{N \times N}$ are computed from the respective $\delta^{m}$ demographic element. The model $f(\cdot)$ to solve the task is given by
\vspace{-0.2cm}
%The objective is to build a model $f(\cdot)$ that learns feature representation based on input affinity graphs, \emph{i.e}~neighbor subjects should have similar feature representation. This yields a smooth label propagation to the unlabeled set $Y_U \in \mathbb{R}^{U \times K}$ as
\begin{equation}
\hat{Y}= f(X, G^{(m)}; \theta).
\end{equation}
The model takes $X$ and $G^{(m)}$ as input to train the parameters $\theta$ and outputs discriminative features for classification. 
Fig. \ref{fig:framework} shows the entire methodology, which can be divided into three main parts: (1) Affinity matrix $W^{m}$ construction, (2) the forward propagation model: we describe the model, where architecture to produce class-separable features and (3) the self-attention layer, for automatic weighting of the graph-specific output features of each branch.\\
%%%%%%% Affinity graph construction
%\textbf{Affinity Graph Construction:}
%Let graph $G^{(m)} = \left \{V, E^{(m)}, W^{(m)} \right \}$ is weighted and undirected consisting of common vertex set $V \in \mathbb{R}^{N}$, each vertex $v_i$ corresponding to each subject $x_i$ respectively and specific edge set $E^{(m)} \in \mathbb{R}^{N \times N}$, and affinity matrix $W^{(m)}\in \mathbb{R}^{N \times N}$. Each reveals distinct intrinsic relationship between the vertices. Edges between vertices are defined based on the given element of demographic data information as 
%\vspace{-0.2cm}
%%We define $E^{(i)}\left ( v,w \right )=\rho \left ( M^{(i)}\left ( v \right ),  M^{(i)}\left ( w \right )\right )$ where $ \rho $ is defined as the Kronecker delta. For quantitative measures such as the subject's age, we define $\rho$ as
%\begin{equation}
%E^{(m)}\left( v_i, v_j\right )=  \begin{cases}
%1 & if \left | M^{m}\left ( v_i \right ) - M^{m}\left ( v_j \right ) \right |< \beta\\
%0 & otherwise
%\end{cases}, 
%\end{equation}
%where $M^{m}(\cdot)$ is the corresponding demographic element, \emph{e.g} gender, age, or location, and $\beta$ is a threshold. A similarity metric between the subjects $Sim(v_i,v_j)$, \emph{e.g.} correlation coefficient, is incorporated to weight the edges as 
%\begin{equation}
%W^{(m)}\left(v_i,v_j\right) =  Sim(v_i,v_j) \circ E^{(m)}\left(v_i,v_j\right),
%\end{equation}
%where $\circ$ is the Hadamard product.\\
\textbf{Affinity Matrix $W^{(m)}$ Construction:}
We construct $M$ affinity matrices corresponding to each of the demographic element. For the $m^{th}$ element, let the graph $G^{(m)} = \left \{X, E^{(m)}\right \}$ be an undirected and unweighted, where all the $M$ graphs have a common vertex set $X$.
%\in \mathbb{R}^{N \times d}$. Each vertex $X_{i}$ bears a $d-$ dimensional feature vector. 
$E^{(m)} \in \mathbb{R}^{N \times N}$ is a demographic element specific edge matrix.
Each graph $G^{(m)}$ reveals distinct intrinsic relationships between the vertices. Edges between vertices are defined based on the given demographic element as 
\vspace{-0.2cm}
%We define $E^{(i)}\left ( v,w \right )=\rho \left ( M^{(i)}\left ( v \right ),  M^{(i)}\left ( w \right )\right )$ where $ \rho $ is defined as the Kronecker delta. For quantitative measures such as the subject's age, we define $\rho$ as
%\begin{equation}
%E^{(m)}\left( v_i, v_j\right )=  \begin{cases}
%1 & if \left | M^{m}\left ( v_i \right ) - M^{m}\left ( v_j \right ) \right |< \beta\\
%0 & otherwise
%\end{cases}, 
%\end{equation}
\begin{equation}
E^{(m)}_{i,j} =  \begin{cases}
1 & if \left | \delta_{i,m} - \delta_{j,m} \right |< \beta\\
0 & otherwise
\end{cases}, 
\end{equation}
where $\delta^{m}(\cdot)$ is the corresponding demographic element and $\beta$ is a threshold. We generate affinity matrix from these graphs by weighting the edges. A similarity metric between the subjects $Sim(X_i,X_j)$, \emph{e.g.} correlation coefficient, is incorporated to weight the edges as 
%\begin{equation}
%W^{(m)}\left(v_i,v_j\right) =  Sim(v_i,v_j) \circ E^{(m)}\left(v_i,v_j\right),
%\end{equation}
\begin{equation}
W^{(m)}_{i,j} =  Sim(X_i,X_j) \circ E^{(m)}_{i,j}\left(X_i,X_j\right),
\end{equation}
where $\circ$ is the Hadamard product.\\
\textbf{Forward propagation model:}
We design our model such that it trains each affinity graph separately. The proposed model bears the parallel setting of $M$ branches as shown in Fig. \ref{fig:framework}. Each branch is equipped with spectral graph theory based GC layers. These layers help to adopt convolutions on graphs unlike grid based convolutions ~\cite{kipf2016semi,defferrard2016convolutional}. The proposed forward propagation model is given by: 
%$H^{(l+1)} = \sigma (\widetilde{D}^{-\frac{1}{2}}\widetilde{W}\widetilde{D}^{-\frac{1}{2}}H^{(l)}\theta ^{(l)})$\\
\begin{equation}
H_{l+1}^{(m)} = \sigma \left ( {D^{(m)}}^{-\frac{1}{2}} W^{(m)} {D^{(m)}}^{-\frac{1}{2}} H_{l}^{(m)} \Theta_{l}^{(m)}\right )
\end{equation}
$D$ is the diagonal matrix with $D_{ii}^{(m)} = \sum_{j} W_{ij}^{(m)}$. $\Theta_{l}^{(m)}$ are the trainable layer-specific filters, which can be derived from a first-order approximation of localized spectral filters on graphs \cite{kipf2016semi}, and $H_{l}^{(m)}$ is the feature representation of the previous layer ($H_{0}^{(m)} = X$). ${D^{(m)}}^{-\frac{1}{2}} W^{(m)} {D^{(m)}}^{-\frac{1}{2}}$ is the normalized graph Laplacian, and $\sigma(\cdot)$ is the rectified linear unit function. The model outputs $H_{logits}\in \mathbb{R}^{N \times K}$. 

\textbf{Self-Attention Layer:}
%The logits $H_{logits}$ are the results of the affinity matrix based on spectral convolutions. Hence 
The logits for $M$ branches differ with respect to each other because of graphs although features on each vertex are common. In order to rank the demographic data elements, we design a linear combination layer that ranks the logits coming from the last hidden layer as%$H^{(m)}_{logits} \in \mathbb{R}^{N \times K}$ as
\vspace{-0.2cm}
\begin{equation}
\hat{Y} = Softmax\left(\sum_{m=1}^{M} \omega_m H^{(m)}_{logits}\right), 
\end{equation}
where $\omega_m$ is the trainable scalar weight associated with the demographic element  %$\mathbb{S}(\cdot)$ is the softmax function, 
and $\hat{Y}$ are the normalized log probabilities. We define our objective function as binary weighted cross entropy loss on the labeled data to train the model parameter.\\ 

\vspace{-0.6cm}
%Singular value decomposition of such graph is used to find out lower dimensional subspace which can preserve the information of the neighborhood within the affinity graph \textcolor{red}{helping in reducing the computation?}\\ 
\section{Experiments}
\label{sec:pagestyle}

%Our experiments have been designed carefully to firstly investigate the influence of individual affinity graph on training as well as inference. Secondly, the second set of experiment deals with multi-layer graph convolutional network with non-trainable weighting layer. Lastly we perform experiment with weights of last layer trainable to investigate the automatic ranking of meta information done by multi- layer GCN.\\
%%%%% RESULTS
%Our experiments have been designed carefully to (1) investigate the influence of individual affinity graphs on the performance of predictive models, (2) to confirm the varied performance over different combination of affinity matrices and (3) to ca to validate our proposed method in the presence of multi-graphs setting compared to the baseline approaches~\cite{parisot2017spectral}. Further, we provide analysis of weighting demographic elements.\\

Our experiments have been designed to (1) investigate the influence of each affinity matrix on the performance of the predictive models, (2) investigate the performance of the predictive model with multi-graph setting approaches~\cite{parisot2017spectral}, (3) we show comparison with 3 methods, linear classifier, two-layered Dense Neural Network, baseline GCN method \cite{parisot2017spectral}, proposed model and (4) investigate in-depth insight of self-attention layer with multi-graph setting.\\
%%%%%
\textbf{Dataset:}
We show results on a publicly available dataset namely Tadpole \cite{marinescu2018tadpole} for the prediction of Alzheimer\textquotesingle s disease. 
The dataset is a subset of ADNI\cite{jack2008alzheimer} consisting of 564 patients. The goal is to classify each patient into one of the three classes Normal, Mild Cognitive Impairment (MCI) and Alzheimer\textquotesingle s disease (AD).
For each patient, the features are collected from various biomarkers (MR, PET imaging, cognitive tests, CSF biomarkers, etc). Further risk factors are provided for each subject in terms of APOE genotyping status and FDG PET imaging. 
%which measures the cell metabolism, where cells affected by AD show reduced metabolism.  
Demographic elements (age and gender) are also provided. Entire data is pre-processed with ADNI's standard data-processing pipeline.%\textcolor{red}{cite}\\ 
\textbf{Implementation:}
Number of features $d$ = 354, dropout rate: $0.3$, $\ell_2$- regularisation: $5 \times 10^{-4}$.
All the experiments are implemented in Tensorflow\footnote{\url{www.tesnsorflow.org}} and performed with Nvidia GeForce GTX 1080 Ti 10 GB GPU.
%Most of the GCN parameters are adopted from the baseline to obtain the fair comparison. The ABIDE network is trained for \textcolor{red}{150 epochs}, %whereas CXR is trained for 500 epochs. 
We use early stopping criteria to decide the number of epochs for each setting. The model is evaluated based on the classification mean accuracy (ACC) for 10-fold Cross-Validation. %and the area under the receiver operating characteristics curve (AUC) on the validation set.0 Further, we report the paired t-test to measure the statistical significance with a significance level of $5\%$. 

\vspace{-0.3cm}
\section{Results and Discussion:}
%width=0.8\linewidth, height=0.35\textheight
\begin{figure*}[htb]
	\setlength{\linewidth}{0.4\textwidth}
		\centering
		\subfigure[Influence of individual affinity with each showing different mean accuracy matrix]{\includegraphics[width=0.8\linewidth]{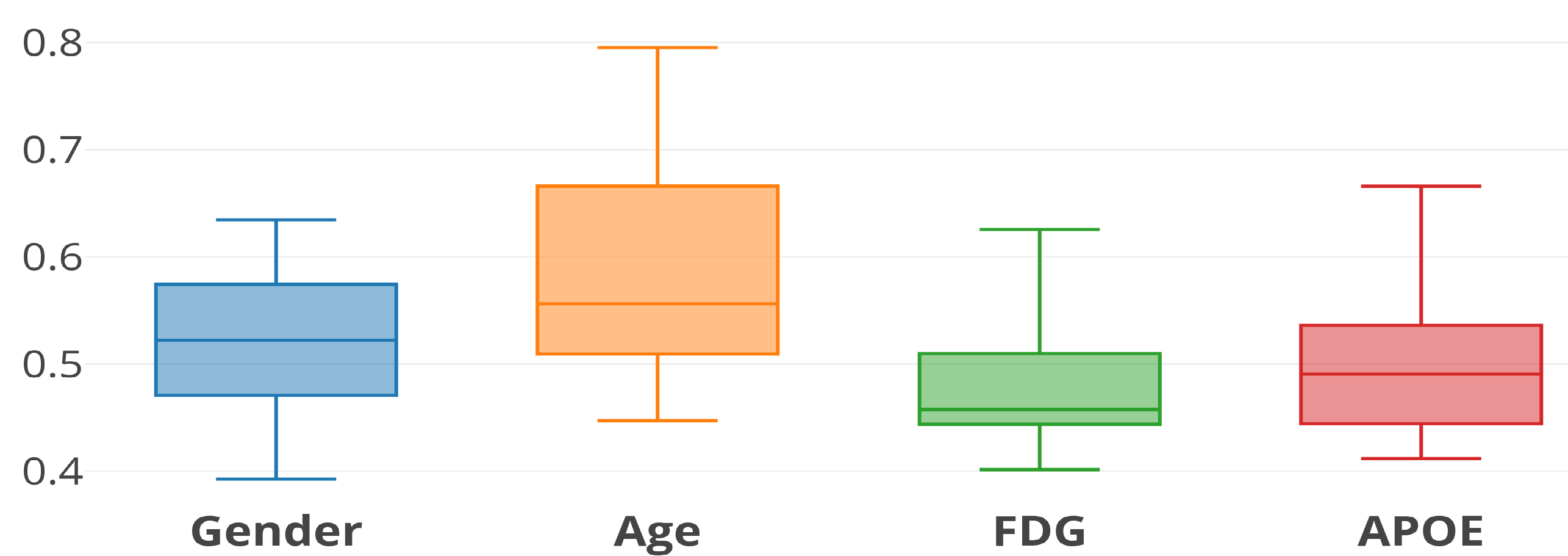}}
		\subfigure[The results over all the comparative methods with proposed model outperforming.]{\includegraphics[width=0.8\linewidth]{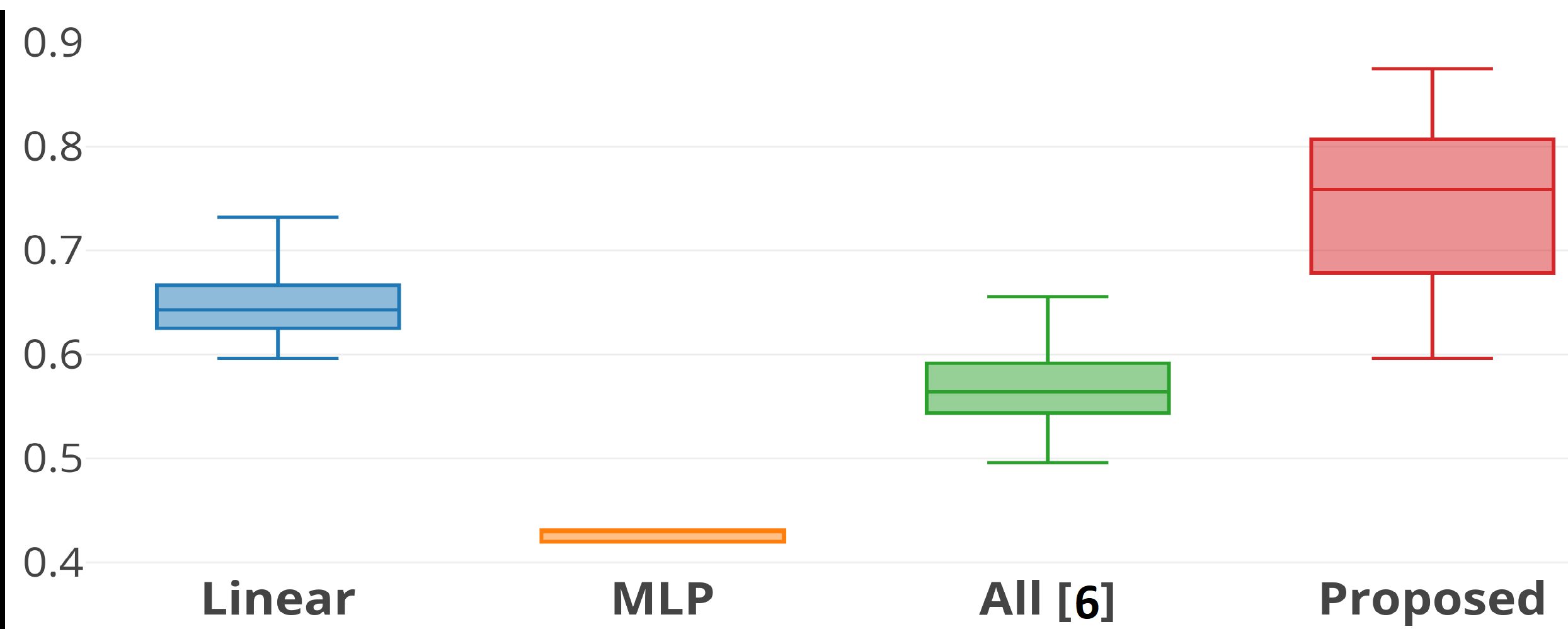}}
		\subfigure[Different combinations of all four affinity matrices showing differnt results]{\includegraphics[width=1.4\linewidth]{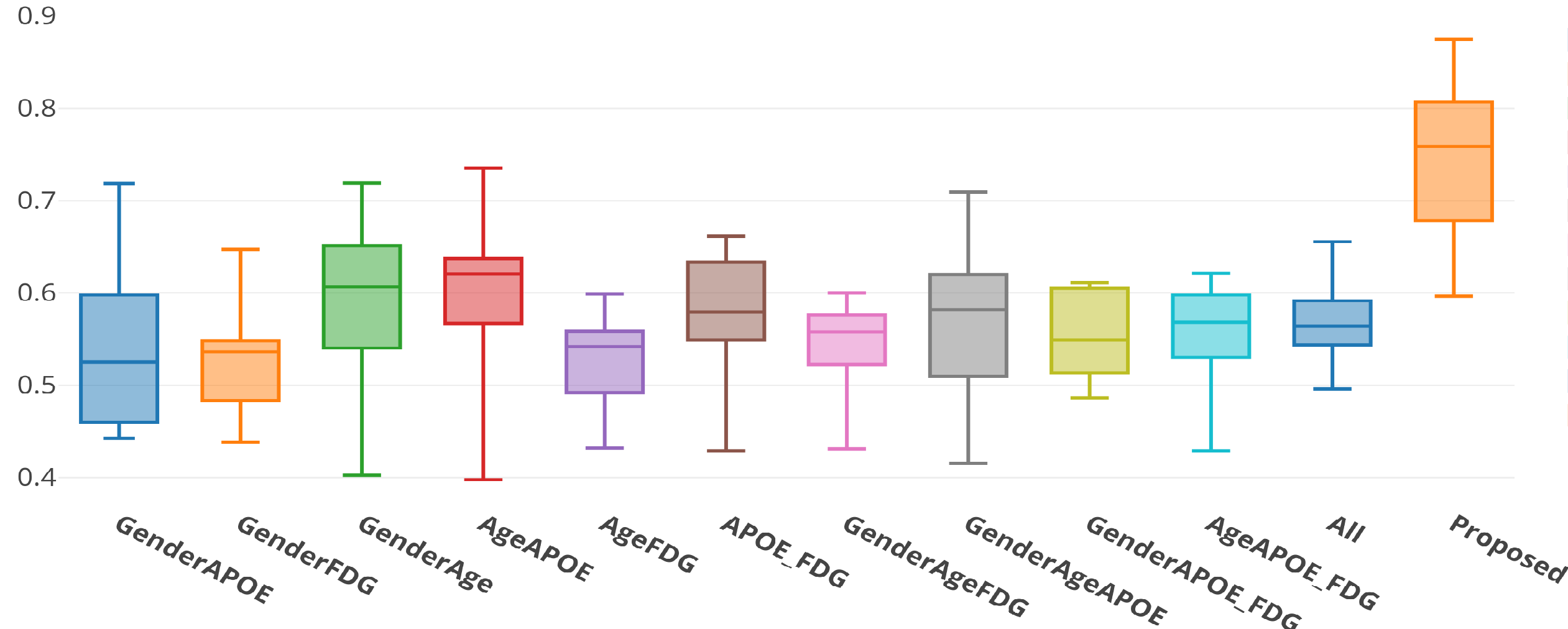}}
	\caption{All the three subfigures show the boxplots of accuracy over 10-folds cross validation.}
	\label{0000-fig-02}
\end{figure*} 

In this section, we discuss the results of all the experiments in detail.\\
\textbf{Influence of individual affinity matrix:} For individual affinities it should be noted from fig. \ref{0000-fig-02} (a) that each graph shows different results. This means that the input affinity matrices have unequal relevance to the task at hand. For example, the age graph shows the best performance and the FDG graph shows the worst. The performance reduces when all the graphs are averaged and used as input as in the baseline method \cite{parisot2017spectral}. This proves that averaging affinity graphs degrade the performance that could have been obtained otherwise.\\
\textbf{Performance with different combinations of graphs:}
We perform another experiment by using all the different combinations of affinity matrices as input. This validates that the performance varies if the combination of affinity matrices is changed. According to \cite{perrin2009multimodal} age and gender are the most important factors compared to APOE and FDG for the prediction of AD. The results are demonstrated in terms of boxplots of accuracies as shown in fig.\ref{0000-fig-02} (c) which confirms the different combination show different result.Moreover, the combination of gender and age show the maximum performance and most of the combinations using FDG and APOE reduce the performance. This depicts that our model upholds the clinical semantic same as \cite{perrin2009multimodal}. This experiment also confirms that the overall performance reduces when all the affinity graphs are weighted equally and averaging deteriorates the positive influence of other affinity matrices due to the loss of neighborhood structure for individual graphs.Our proposed model with self-attention outperforms all the combinations, since it captures the correct weighting required for optimal performance.\\
\textbf{Performance in comparison to other methods:}
We compare the proposed method to three state-of-the-art methods namely linear classifier, neural network and \cite{parisot2017spectral} as shown in fig. \ref{0000-fig-02} (b). We chose these methods respectively to investigate 1) how linearly separable the features at every node are? 2) what is the performance of the model when features are concatenated? 3) what is the significance of incorporating the graph for the task? and 4) how important is it to weight the graphs? From fig. \ref{0000-fig-02} (b), it can be seen that features are separable as the linear classifier performs quite well compared to two other methods shown. For NN, where the features are concatenated the model architecture becomes the problem. We used the same number of hidden layers (2) and hidden unties (16, 3 respectively) with the input of the feature dimension of 354. NN fails to perform well with this architecture. As can be seen that the baseline \cite{parisot2017spectral} improves the performance with respect to NN showing the strength of the GCN, however, it performs lower than linear and proposed. This is due to the corrupted combination of the neighborhood. Finally, our proposed method outperforms all the methods with the correct weighted combination of neighborhood and $H_{logits}$.\\
%\par The proposed model with self-attention incorporated, shows significantly better performance than individual affinities as well as average graph. In case of Autism, the patient is born with Autism, hence for ABIDE $W^{g}$ or $W^{s}$ both do not play much important role in the prediction of Autism. This is reflected in the results from table.~\ref{tab:abide}.\\
\textbf{Effect of self attention: }We also investigated the weights learnt for each branch by our model. The self-attention layer learned maximum weight for gender and age (0.35 and 0.27 respectively) and lower weight for FDG and APOE(0.09 and 0.29 respectively). It is confirmed from \cite{perrin2009multimodal} that age and gender are significant factor for predicting AD.
%\par Over all from all the datasets we can conclude that:
%1) Neighborhood definition is crucially for spectral convolution.
%2) Their relevance is different for different dataset.
%3) our model works best for dataset with large no. of affinity matrices.
\vspace{-0.4cm}
\section{Conclusion}
\label{sec:typestyle}
\vspace{-0.2cm} 
All our experiments go inline with our hypothesis that affinity graphs influence the performance of disease prediction differently. GCNs are sensitive to the defined neighborhood. Combination of affinities alters the possible neighborhood between the subjects. Further, our proposed method with self-attention clearly incorporates the unequal contributions of graphs and outperforms all the setups with significant margin. The order of complexity for our model versus the baseline model \cite{parisot2017spectral} is nearly equal as $O(n)\approx O(2n)$, making it scalable for a larger number of demographic elements. We train the GC layers first for 150 epochs and then let the self-attention layer train further. This helps channelize the learning of weights of GC layers as well as self-attention layer. The features at every node are kept simpler to gain more insights about effect og graphs.
\par Further the choice of thresholds for creating the graphs are followed from clinical statistics provided by the literature. One might argue that splitting a single graph into multiple graphs will decrease the performance as some connections are lost in the thresholding process. However aggregating the graphs from different information source will lead to the loss of individual structure and unequal relevance cannot be considered.
%As a future work 1) out of sample extension is required for this method, which will boost its usage for the live datasets, 2) for improving the performance weight of every node can also be investigated, 3) better features can be extracted for every node, as in this paper they are kept simple to investigate the power of graph. 

% Below is an example of how to insert images. Delete the ``\vspace'' line,
% uncomment the preceding line ``\centerline...'' and replace ``imageX.ps''
% with a suitable PostScript file name.
% -------------------------------------------------------------------------

%\vfill
%\pagebreak

\vspace{-0.4cm}
\bibliographystyle{IEEEbib}
\bibliography{ISBI2019_ISBItemplate_biblio_new}

\end{document}